\documentclass[letterpaper, 10 pt, conference]{ieeeconf}
\IEEEoverridecommandlockouts   % needed because we use \thanks
\usepackage{amsmath,amssymb}
\usepackage{algorithm}
\usepackage{algpseudocode}

\usepackage{graphicx}
\graphicspath{{../}}
\usepackage{subcaption}
\usepackage{booktabs}
\usepackage{array}
\usepackage{multirow}
\usepackage{wrapfig}

\usepackage[utf8]{inputenc}
\usepackage[dvipsnames,table]{xcolor}
\newcommand{\pgcell}[1]{\cellcolor{ForestGreen!30}#1}
\usepackage{hyperref}

\title{\LARGE \bf ActionGround: Training-Free Runtime Refinement of Frozen VLA Policies}

\author{
Namai Chandra\textsuperscript{1},
Madhur Thareja\textsuperscript{2,1},
Shriram Damodaran\textsuperscript{3},
Addison Lin Wang\textsuperscript{3}
\\
\textsuperscript{1}Indian Institute of Technology Madras
\\
\textsuperscript{2}Indian AI Research Organisation
\\
\textsuperscript{3}Nanyang Technological University
}

\begin{document}
\maketitle
\thispagestyle{empty}
\pagestyle{empty}

%==============================================================================
\begin{abstract}
Vision-Language-Action (VLA) models map visual observations and
language instructions directly to robot actions through a single
end-to-end neural policy, but nothing in that policy explicitly
represents the discrete phase structure of a manipulation task or the
rigid-body dynamics the arm must obey while executing it. We present
\textbf{ActionGround}, a \textbf{neuro-symbolic, training-free runtime
layer} that wraps a frozen VLA policy without retraining,
fine-tuning, or weight access, adding less than $1$\,ms of overhead
per control step. A \emph{symbolic} phase-aware finite-state machine
reads the scene state to identify which manipulation phase (approach,
grasp, transport, place) the episode is in and applies a
phase-appropriate rule-based correction; in parallel, an
\emph{always-on, inertia-weighted} Euler-Lagrange term folds the
robot's equations of motion into every control step, using the
dynamics residual as a logged consistency diagnostic rather than a
gate. Neither channel touches the VLA's weights. Evaluated across
OpenVLA, OpenVLA-OFT, Force-VLA, and Generalist-VLA on ten
LIBERO-Spatial pick-and-place tasks with a 7-DoF Franka Panda, the
same fixed-parameter framework (no per-task or per-backbone retuning)
delivers absolute success-rate increases of up to $6\%$ and absolute
stability increases of up to $19\%$ across all four backbones
(Table~\ref{tab:gap}), with trajectory-efficiency gains of up to
$15\%$ and per-task regressions confined to two named OpenVLA tasks
(Sec.~\ref{sec:results}); it also shows roughly a $10{\times}$
improvement in trajectory-jerk robustness under injected action noise
on a Robosuite cross-simulator sweep. We also ran a matched-seed
Robosuite simulation companion to a real Agilex Piper pick-and-place
trial under an identical injected-bias protocol, lifting Baseline
success from $35\%$ to $95\%$. The physical-hardware trial itself is
a separate, qualitative deployment demonstration; per-trial
quantitative success on the real arm is left to future work. We
evaluate exclusively on rigid-object pick-and-place manipulation and
scope our claims to that setting.
\end{abstract}

\smallskip\noindent\textit{Index Terms}---neuro-symbolic robotics,
vision-language-action models, finite-state phase correction,
Euler-Lagrange consistency diagnostic, frozen-policy adaptation.

%==============================================================================
\section{Introduction}

The ambition to build robotic agents that perceive, reason, and
physically interact with the world in response to language
instructions has long been a central objective in embodied
AI~\cite{kaelbling2020foundation}. Real-world manipulation is
inherently multi-physical: a robot must respect rigid-body dynamics, contact
interactions, friction, and gravity while interpreting visual scenes and
following linguistic commands.

\begin{figure}[t]
    \centering
    \includegraphics[width=0.9\linewidth]{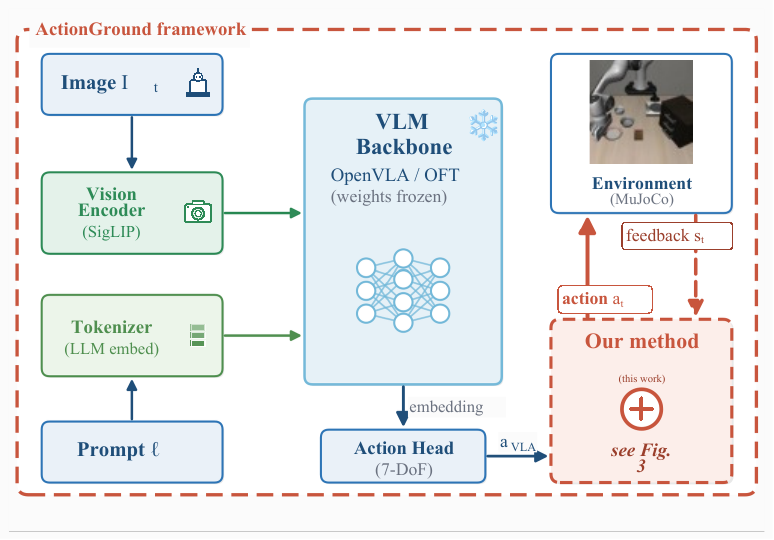}
    \caption{\textbf{Overview of our ActionGround framework}. See main texts for details.}
    \label{fig:vla-pipeline}
\end{figure}

Recent vision-language-action (VLA) models
provide an end-to-end mapping from observations and language
prompts to actions (Fig.~\ref{fig:vla-pipeline}). Systems such as
RT-2~\cite{brohan2023rt2}, $\pi_0$~\cite{black2024pi0},
CogACT~\cite{li2024cogact}, TinyVLA~\cite{wen2025tinyvla}, and
SmolVLA~\cite{shukor2025smolvla} support language-conditioned control across diverse embodiments.
However, these architectures are trained purely to fit demonstration data
and do not explicitly encode the equations of motion or contact constraints
that govern physically feasible trajectories. As a result, current VLA
models remain limited in structured physical reasoning and long-horizon
embodied understanding~\cite{liu2023libero,kim2024openvla,kim2025openvlaoft}.

Evaluated on the LIBERO-Spatial benchmark~\cite{liu2023libero}, our empirical analysis reveals that a single-step OpenVLA policy attains a
success rate of only $36\%$, while its chunked, memory-augmented variant
OpenVLA-OFT reaches $92\%$. This gain shows that short-horizon temporal
coherence mitigates many errors, but also exposes a structural limitation:
even with chunking, the policy reasons only over a few recent steps and
lacks a global view of the manipulation sequence, leading to \textbf{limited
task-level physical coherence} in multi-phase, contact-rich manipulation,
and higher computational and memory cost than lighter single-step VLA models.
\begin{figure}[t]
    \centering
    \includegraphics[width=0.9\linewidth]{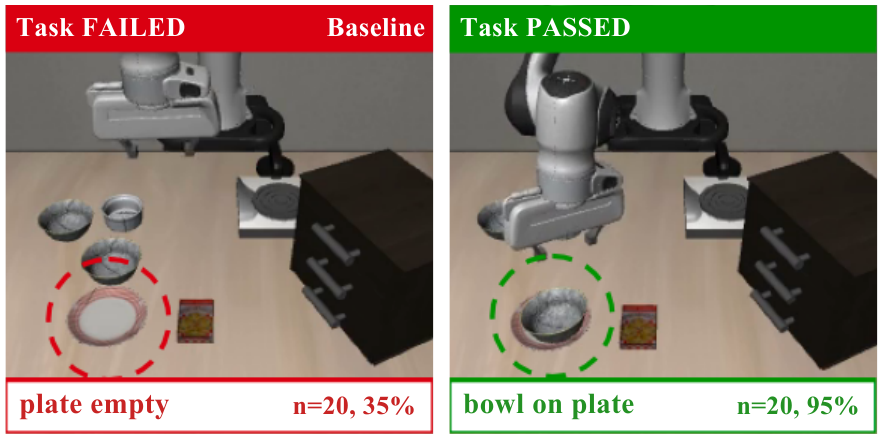}
    \caption{\textbf{OpenVLA backbone, same weights, on the
    matched-seed Robosuite place-on-plate simulation companion
    described in Sec.~\ref{sec:results} ($n{=}20$ per mode).}
    ActionGround succeeds where the unmodified backbone fails under
    the injected placement-phase bias.}
    \label{fig:hero-inline}
\end{figure}

Rather than integrate physics into the policy at training time or
redesign it~\cite{raissi2019pinn,lutter2019deep,cranmer2020lagrangian_nn,greydanus2019hamiltonian,zhong2020symoden,dawson2023survey},
we consider the setting in which an already-trained VLA must be improved while keeping its weights and interface fixed.
ActionGround is a \textbf{training-free, modular, plug-and-play layer}
that operates alongside an existing policy, rectifying actions
immediately prior to execution through any frozen VLA backbone's
standard 7-DoF action vector, with no parameter retraining,
fine-tuning, or internal weight access, while improving task success on three backbones and trajectory quality across all four (Sec.~\ref{sec:results}).
ActionGround reads the environmental state and applies phase-conditioned
rather than uniform corrections, with the dynamics residual logged as
a continuous consistency diagnostic. Concretely, ActionGround forms a
discrete--continuous runtime interface: a symbolic task state selects
the applicable correction law, and physical information constrains how
the proposed action is refined, combining the learned policy with discrete task structure and analytical dynamics at runtime, without
modifying the policy itself. Sec.~\ref{sec:branchA} details the
two-branch design (a symbolic phase-aware finite-state machine and an
always-on, inertia-weighted Euler-Lagrange correction, together adding
$<\!1$\,ms of latency per control step).
This paper makes three major contributions: \textbf{(I)}
We provide an empirical
analysis of the physics gap in both single-step and memory-augmented VLA models. 
% Results reveal that while naive temporal averaging improves trajectory smoothness, it introduces a severe trade-off by reducing overall task success. 
\textbf{(II)} We propose ActionGround, a novel plug-and-play, training-free neuro-symbolic runtime layer for frozen VLA backbones with a 7-DoF action interface. \textbf{(III)} We evaluate four distinct VLA paradigms
  (single-step, chunked, force-conditioned, and generalist) alongside
  four inference modes under a unified PyTorch stack. Our results
  demonstrate consistent improvements in task success, precision, and
  efficiency, with mean trajectory jerk held to $+17\%$ across a
  Robosuite noise sweep where the Baseline degrades by $+175\%$,
  and no per-task regressions on Force-VLA or Generalist-VLA. We additionally
  demonstrate the same corrector on an Agilex Piper real-world
  pick-and-place as a qualitative deployment study.
% The framework integrates a
% phase-aware finite-state machine with a selective Euler-Lagrange gate
% to conditionally enforce geometric and dynamical constraints without
% modifying the underlying model weights. 

% \begin{figure}[t!]
%     \centering
%     \includegraphics[width=\linewidth]{figures/fig_vla_pipeline.pdf}
%%     \caption{\textbf{Overview of our ActionGround framework}. See main texts for details.}    
%     % The \emph{only} language input is prompt $\ell$; the
%     % on-board RGB image $I_t$ is taken by the VLA's vision encoder. The
%     % frozen VLA backbone (OpenVLA, OpenVLA-OFT, Force-VLA, or Generalist-VLA)
%     % emits a raw 7-DoF action $a_{\mathrm{VLA}}$ which the post-hoc
%     % \textbf{ActionGround} framework intercepts before actuation. ActionGround reads the
%     % MuJoCo state $s_t$ \emph{only} and runs two channels, a phase-aware FSM
%     % (Branch~A) and a selective Euler-Lagrange gate (Branch~B). No VLA
%     % weights are updated and the framework adds $<\!1$\,ms per control step.}
%     \label{fig:vla-pipeline}
%% \end{figure}

%==============================================================================
\section{Related Work}

\noindent \textbf{Neuro-Symbolic Integration in Robotics.}
A parallel line of work builds discrete symbolic structure directly
into robot decision-making rather than leaving it implicit in a single
network. Task-and-motion-planning frameworks compose symbolic
operators over continuous state with sampling-based motion
planners~\cite{garrett2020pddlstream,kaelbling2020foundation}, and such
operators can themselves be learned from continuous skill execution
rather than hand-specified~\cite{konidaris2018skills,silver2021learning}.
Neuro-symbolic AI more broadly studies how discrete symbolic
representations and continuous neural computation inform each other,
from probabilistic-logic layers~\cite{manhaeve2018deepproblog} and
differentiable logic~\cite{riegel2020lnn} to concept learners that
ground symbols directly in perception~\cite{mao2019nscl}; see
~\cite{garcez2020nesy} for a survey of this ``third wave.''
ActionGround adopts the same discrete/continuous split as this
literature, but couples the two directions explicitly at runtime
rather than through joint training: the symbolic phase $\phi_t$
selects which correction law is active (Eq.~\ref{eq:phase-detect}),
while an always-on, inertia-weighted Euler-Lagrange correction
(Eq.~\ref{eq:dyn-blend}) refines that phase-selected action on every
step, per-joint, by how dynamically loaded the current configuration
is; the Euler-Lagrange residual itself is logged as a consistency
diagnostic and does not modulate this weighting. Unlike TAMP, where
the symbolic layer plans and the continuous layer merely samples
feasible motions, here the continuous physical signal also refines the
symbolic one's output, a two-way runtime interaction without ever
updating $\pi_\theta$'s weights.

\noindent \textbf{Vision-Language-Action (VLA) Models.}
VLA models have become the standard paradigm
for language-conditioned manipulation. Early systems such as
RT-1/RT-2~\cite{brohan2022rt1,brohan2023rt2} and PaLM-E~\cite{driess2023palme}
showed that large vision--language backbones can be adapted into end-to-end
robot controllers, and subsequent works have diversified the action-head
family: autoregressive decoders and policy distillation in
$\pi_0$~\cite{black2024pi0} and CogACT~\cite{li2024cogact}, diffusion-based
heads in Diffusion-VLA~\cite{wen2024diffusionvla}, and more efficient
variants such as RT-H~\cite{belkhale2024rth}, TinyVLA, and
SmolVLA~\cite{wen2025tinyvla,shukor2025smolvla} for deployment. Beyond
single-arm settings, hierarchical planners~\cite{ahn2022saycan,huang2023voxposer,huang2022innermonologue},
multi-embodiment controllers~\cite{shridhar2022cliport,stone2023moo,zhao2023act,openx2024},
and recent surveys~\cite{sapkota2025vla,li2024robovlms}
document this rapid progress. Across these works, however, \emph{physical
structure remains implicit: architectures are trained to fit demonstration
data, but no component explicitly enforces equations of motion, contact
feasibility, or energy balance.} OpenVLA and its optimised fine-tune
OpenVLA-OFT~\cite{kim2024openvla,kim2025openvlaoft} highlight this gap:
chunked inference improves success substantially on LIBERO-Spatial, showing
the value of short-horizon temporal coherence, yet residual failures
persist on contact-rich tasks, indicating a remaining physics
deficit~\cite{liu2023libero,zhang2025vlabench}.

\noindent \textbf{Physics-Informed Learning and Control.}
Another line of research incorporates analytical dynamics
into learning and control. Recent techniques include \emph{soft residual}
penalties, adding equation-of-motion terms to the training loss as in
PINNs~\cite{raissi2019pinn}. A complementary approach is to adopt
\emph{hard structural} constraints, embedding Lagrangian, Hamiltonian, or
symplectic structure directly into the network~\cite{lutter2019deep,cranmer2020lagrangian_nn,greydanus2019hamiltonian,zhong2020symoden},
often coupled with differentiable simulators~\cite{degrave2019diffsim,toussaint2018differentiable,freeman2021brax,heiden2021neuralsim}
or graph-based dynamics models~\cite{sanchez2018graph_nets_physics,li2019propagation_networks}.
More recent works have focused on learning-based techniques such as
\emph{stability-certified} controllers, including Lyapunov-based RL and
control barrier functions~\cite{chow2019lyapunov,achiam2017cpo,chang2019neural,ames2017cbf,dawson2023survey},
as well as MPPI~\cite{williams2017mppi}, DMPs~\cite{ijspeert2013dmp},
RMPs~\cite{ratliff2018rmp}, and OSC~\cite{khatib1987osc}. These approaches
demonstrate that encoding physics can improve sample efficiency and
robustness, but they typically act during training, assume access to
gradients through the dynamics, or require redesigning the controller.
\emph{ActionGround instead treats physics as an inference-time correction applied
to a frozen VLA: the backbone and its training remain unchanged, and
dynamics enter only through an always-on runtime injector}
(Table~\ref{tab:constraint-spectrum}). Unlike CBF-QP safety
filters~\cite{ames2017cbf,dawson2023survey} which solve a per-step
quadratic program against an always-on certificate, or MPPI-style
refinement~\cite{williams2017mppi} which samples and re-scores
rollouts against a learned cost, ActionGround evaluates a single
closed-form residual and applies an always-on, inertia-weighted
correction every step, so it adds $<\!1$\,ms per step
without an optimiser in the loop. Likewise, recent force-aware
reactive VLAs (ForceVLA / FD-VLA-style heads) achieve contact
correction but require training-time access to a force-residual
head; ActionGround targets the same failure modes purely at inference,
with no weight updates.

\begin{table*}[t]
\caption{Taxonomy of physics-constraint strategies in robotic policy learning.
``Runtime'' indicates whether the constraint is active at inference.}
\label{tab:constraint-spectrum}
\centering
\small
\begin{tabular}{@{}llcc@{}}
\toprule
\textbf{Strategy} & \textbf{Representative works} & \textbf{Guarantee} & \textbf{Runtime} \\
\midrule
Soft Residual
    & PINNs~\cite{raissi2019pinn}, PIPER~\cite{chandra2026piper}
    & Approximate & No \\
Hard Structural
    & DeLaN~\cite{lutter2019deep}, HNNs~\cite{greydanus2019hamiltonian},
      SymODEN~\cite{zhong2020symoden}
    & Exact & No \\
Stability-Certified
    & Neural Lyapunov~\cite{chang2019neural}, CBF~\cite{dawson2023survey}
    & Certified & Optional \\
\textbf{ActionGround (ours, runtime)}
    & FSM + EL residual (this work)
    & Diagnostic & \textbf{Yes} \\
\bottomrule
\end{tabular}
\end{table*}

\noindent \textbf{World Models, Diffusion Policies, and Inference-Time Adaptation.}
Latent world models, diffusion policies, and constraint-grounded
representation
learning~\cite{ha2018worldmodels,hafner2023dreamerv3,chi2023diffusionpolicy,nair2023r3m,ma2024eureka}
improve long-horizon behaviour and perception but remain
physics-agnostic in the sense above. Closer in spirit to our setting
are inference-time interventions that modify actions without retraining the
policy. Temporal ensembling in ACT and Octo~\cite{zhao2023act,ghosh2024octo}
applies a uniform exponential moving average over position dimensions,
improving smoothness but ignoring task phase and dynamics. In our
experiments, this kind of uniform smoothing reduces OpenVLA success on
LIBERO-Spatial by flattening responsive motions during critical contact
phases~\cite{liu2023libero}. \emph{ActionGround can be viewed as a structured
alternative: it conditions corrections on the manipulation phase and
applies an always-on, inertia-weighted dynamics correction on top,
logging the residual as a diagnostic rather than using it to switch
physics on or off.}

% \noindent \textbf{Benchmarks for Manipulation.}
% We evaluate primarily on LIBERO-Spatial~\cite{liu2023libero}, a suite of
% spatial reasoning manipulation tasks that combine nontrivial object
% geometry with contact-rich phases, making them sensitive to both temporal
% coherence and physical consistency. Other benchmarks such as
% CALVIN~\cite{mees2022calvin}, ManiSkill~\cite{mu2021maniskill},
% RLBench~\cite{james2020rlbench}, Meta-World~\cite{yu2020metaworld},
% ALFRED~\cite{shridhar2020alfred}, and SIMPLER~\cite{li2024simpler} target
% complementary aspects of manipulation and embodied reasoning. \emph{Our
% focus on LIBERO-Spatial is deliberate: it exposes the gap between
% short-horizon temporal fixes (e.g., chunking and smoothing) and the
% remaining physics errors that ActionGround is designed to address.}

%==============================================================================
\section{Methodology}
\subsection{Basic Principle of VLA}

A vision-language-action (VLA) policy $\pi_\theta$ is a learned function
that regresses a low-level robot action from an on-board image $I_t$ and a
natural-language prompt $\ell$:
\begin{equation}
    a_{\text{VLA}} = \pi_\theta(I_t, \ell) \in \mathbb{R}^7,
    \label{eq:vla-output}
\end{equation}
\begin{equation}
    a_{\text{VLA}} = [\Delta x, \Delta y, \Delta z, \Delta\phi, \Delta\theta,
    \Delta\psi, g]^\top,
    \label{eq:vla-output-components}
\end{equation}
where the first six components specify a delta end-effector pose and the
seventh is a gripper command. The policy is trained purely on demonstration
data and exposes no built-in constraint on the robot's equations of motion
or on the phase-specific physics of manipulation: $\pi_\theta$ has no
notion of which manipulation phase the trajectory is currently in, and no
check that the proposed $a_{\text{VLA}}$ is kinodynamically consistent
with the joint state $(q_t,\dot{q}_t)$ before actuation.

\subsection{Empirical Results: Identifying the Physics Gap}

This missing structure leads to two recurring failure modes in our experiments: $\pi_\theta$ is free to emit actions that the
underlying robot dynamics will resist, distort, or refuse. Two qualitative
failure modes recur across every backbone we evaluated. First, the policy
issues lateral motion at the moment of contact (the ``grasp'' phase) even
when the gripper is not over the target object, producing premature closes
or empty-hand grasps. Second, in the placement phase the policy enters at
near-full velocity without a deceleration profile, overshooting
sub-centimetre targets. Off-the-shelf inference-time fixes do not address these failure modes: a uniform exponential moving average (EMA) over the action
stream improves trajectory smoothness on the macro scale but flattens the
responsive bursts the policy needs during contact, trading task success
for stability. 

\begin{table}[t]
    \centering
    \footnotesize
    \setlength{\tabcolsep}{4pt}
    \caption{ActionGround gain over Baseline on LIBERO-Spatial
    (aggregate, \%).}
    \label{tab:gap}
    \begin{tabular}{@{}l c c@{}}
        \toprule
        Backbone & $\Delta$\,Stab & $\Delta$\,Succ \\
        \midrule
        OpenVLA                 & $+16.7\%$ & $+2\%$ \\
        Force-VLA               & $+18.2\%$ & $+4\%$ \\
        Generalist-VLA          & $+19.3\%$ & $+6\%$ \\
        OpenVLA-OFT$^{\dagger}$ & $+2.8\%$  & $+0\%$  \\
        \bottomrule
    \end{tabular}
    \par\smallskip\scriptsize $^{\dagger}$chunked decoding;
    others are single-step.
\end{table}
Memory-augmented variants (e.g.\ chunked decoding) recover
local temporal coherence but still reason only over a short window and
miss the phase-level physical context. Smoother action streams alone do not address these failure modes; it requires phase-aware, dynamics-aware corrections
that fire only when the policy is about to commit a physically
inconsistent action.

Table~\ref{tab:gap} backs this empirical claim at the aggregate
level: 
Baseline success on LIBERO-Spatial clusters in a narrow
$36$-$40\%$ band across the three single-step backbones (OpenVLA,
Force-VLA, Generalist-VLA) and uniform temporal smoothing
\emph{degrades} every one of them, confirming that smoothing alone
does not close the gap. The chunked OpenVLA-OFT lands at $92\%$ but
at significantly higher inference cost, and still loses on
contact-rich tasks (e.g.\ T5 at $40\%$). These results indicate limitations in both temporal smoothing and single-step inference, motivating the phase-aware, dynamics-aware design introduced next.

% \begin{table}[h]
%     \centering
%     \footnotesize
%     \setlength{\tabcolsep}{6pt}
%     \caption{Aggregate LIBERO-Spatial success rate (\%) under the
%     Baseline frozen backbone and under uniform temporal smoothing.
%     Temporal smoothing trades success for stability on every
%     single-step backbone, motivating the phase-aware design.
%     Full per-task numbers in
%     Table~A2 of the supplementary.}
%     \label{tab:gap}
%     \begin{tabular}{l c c c}
%         \toprule
%         Backbone & Type & Baseline\,(\%) & Temporal\,(\%) \\
%         \midrule
%         OpenVLA          & single-step & 36 & 28 \\
%         Force-VLA        & single-step & 40 & 36 \\
%         Generalist-VLA   & single-step & 36 & 26 \\
%         OpenVLA-OFT      & chunked     & 92 & 92 \\
%         \bottomrule
%     \end{tabular}
% \end{table}

\subsection{ActionGround Design}
\begin{figure*}[t]
    \centering
    \includegraphics[width=0.85\textwidth]{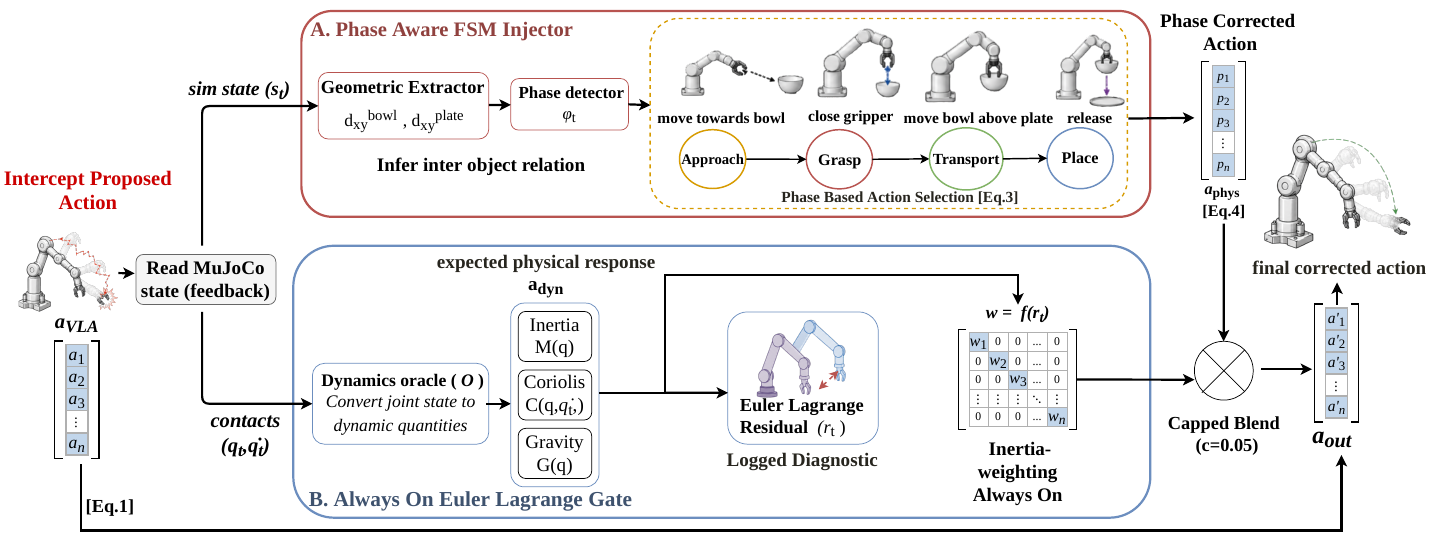}
    \caption{\textbf{ActionGround's runtime refinement layer.} Channel A
    (symbolic phase grounding) and Channel B (always-on, inertia-weighted
    Euler--Lagrange correction) each propose a candidate correction; the
    two are combined and capped-blended with the frozen VLA's own action.
    The Euler--Lagrange residual $r_{\text{EL}}$ (dashed branch) is a
    logged diagnostic only and does not gate or scale the executed
    correction. See Sec.~III-C for details.}
    \label{fig:corrector}
\end{figure*}

We address these errors without retraining $\pi_\theta$. ActionGround is a
\textbf{two-branch corrector} that sits between the VLA's predicted
action $a_{\text{VLA}}$ and the simulator. At every control step,
the MuJoCo state
$s_t = (p^{\text{eef}}_t,\, q^{\text{grip}}_t,\, p^{\text{obj}}_t,\,
\mathbf{c}_t)$ feeds into two independent branches running in
parallel (Figure~\ref{fig:corrector}):

\noindent\textbf{Branch~A} is a phase-aware finite-state machine.
It looks at $s_t$ and decides which manipulation phase the robot is
currently in (approach, grasp, transport, or placement), then applies
a small, phase-specific tweak to the action. The correction needed near contact (slow down, reduce gripper bias) differs from the correction used in free space (smooth jitter, maintain motion). Branch~A encodes that intuition with a rule per phase,
not as a learned weight.

\noindent\textbf{Branch~B} is an always-on, inertia-weighted Lagrangian
diagnostic and blend, applied \emph{on top of} Branch~A's proposal
when active. It computes the analytical residual $r_{\text{EL}}$ of
the robot's equations of motion for the proposed action as a
continuous kinodynamic-consistency signal, and folds a mass-aware
correction into \emph{every} step: joints with lower generalised
inertia, which carry the greater kinetic risk from an inconsistent
command, receive proportionally more correction. We also tested a
hard variant that gates this correction on $\|r_{\text{EL}}\|$
exceeding a threshold $\epsilon$ and found no consistent task gain
over the always-on blend,
so whenever Branch~B is active, every number reported uses the
always-on blend described below rather than the gated variant. The
canonical per-backbone numbers in Table~\ref{tab:pertask}
use \emph{Branch~A only}, uniformly across all four
backbones; Branch~B is reported separately, as an OpenVLA-only
addition on top of Branch~A (Sec.~\ref{sec:results}), because
OpenVLA-OFT's action chunking already supplies comparable step-level
coherence. Branch~B therefore provides an additional correction for errors that Branch~A cannot identify from geometry alone.

When Branch~B is active, its output is combined with the Branch~A proposal through Eq.~\eqref{eq:dyn-blend}: Branch~A
and Branch~B's always-on inertia-weighted correction blend internally
via Eq.~\eqref{eq:dyn-blend} to produce a single physics candidate
$a^{\text{phys}}_t$, which is then combined with $a_{\text{VLA}}$
through the global capped blender
\begin{equation}
    a_t = (1{-}c)\,a_{\text{VLA}} + c\,a_{\text{phys}},\quad c = 0.05,
    \label{eq:blend}
\end{equation}
with a hard gripper override at $|g| > 1.5$ so deliberate grasp
commands are never suppressed by the cap. The executed action is
mostly ($95\%$) the VLA's own prediction, refined by a small ($5\%$)
physics correction: The small correction can accumulate over multiple steps while still allowing the VLA action to dominate an individual control step. The cap, hyperparameter-free across all four
backbones and all 10 LIBERO-Spatial tasks, is the only place
ActionGround can touch the trajectory; no VLA weight is ever updated,
and the full pipeline adds $<\!1$\,ms per step. The rest
of this section details what each branch does and why.

\paragraph{Branch A: Phase-Aware Finite-State Machine Injector}
\label{sec:branchA}

A fundamental structural observation is that different manipulation phases are governed by
qualitatively different physical constraints. A VLA policy applying uniform corrections
across the full episode will necessarily be mismatched to phases it was not optimized for,
exactly the failure of temporal smoothing baselines. Our first corrector channel resolves
this through a rule-based finite-state machine (FSM) that partitions each episode
into phases using geometric predicates on $s_t$, then applies a phase-specific correction:
\begin{equation}
    \resizebox{\linewidth}{!}{$\displaystyle
    \phi_t = \textsc{DetectPhase}(s_t) =
    \begin{cases}
        \texttt{approach}   & d^{\text{bowl}}_{xy} \geq 6\,\text{cm} \\
        \texttt{grasp}      & d^{\text{bowl}}_{xy} < 6\,\text{cm},\;
                              p^{\text{bowl},z}_t < z_{\text{lift}} \\
        \texttt{transport}  & p^{\text{bowl},z}_t \geq z_{\text{lift}},\;
                              d^{\text{plate}}_{xy} \geq 6\,\text{cm} \\
        \texttt{place}      & p^{\text{bowl},z}_t \geq z_{\text{lift}},\;
                              d^{\text{plate}}_{xy} < 6\,\text{cm}
    \end{cases}$}
    \label{eq:phase-detect}
\end{equation}
where $d^{\text{bowl}}_{xy}$ and $d^{\text{plate}}_{xy}$ are horizontal distances from
the end-effector to the bowl and plate, and $z_{\text{lift}} = 1\,\text{cm}$
above the table top (LIBERO-Spatial coordinate frame) is the lift
threshold, shared across all backbones and tasks. Given $\phi_t$, the
phase correction $\textsc{PhaseCorrect}$ modifies $a^{\text{raw}}_t$ as follows.
The predicate uses only the instantaneous geometric state: it carries no
memory of the previous phase and no contact sensing, so it cannot
recover once a height-based threshold is wrong for a given task's
geometry. Tasks whose target starts already elevated relative to the
tabletop (T4, T9) are exactly where this shows up, consistent with the
occlusion-driven phase-detection failure noted in the Limitations
section.

\smallskip\noindent\textit{Approach.}  A premature-grasp veto sets
$g\!\leftarrow\!0$ whenever $d^{\text{bowl}}_{xy} \geq \delta_{\text{grasp}}$
($\delta_{\text{grasp}} = 6$\,cm, the same threshold as the phase
predicate, so the veto is active for the entire \texttt{approach}
phase), enforcing the geometric precondition for contact, which VLAs
routinely ignore.

\smallskip\noindent\textit{Grasp.}  A guidance bias $\beta = 0.5$ blends
$a^{\text{raw}}_t$ toward the computed grasp waypoint $p^*$,
$a^{\text{phys}}_t = \beta(p^* - p^{\text{eef}}_t) + (1-\beta)\,a^{\text{raw}}_t$.
No smoothing is applied, high-frequency responsiveness is essential at
contact acquisition.

\smallskip\noindent\textit{Transport.}  A vertical lift bias
$\Delta z^+ = +2\,\text{cm}$ counteracts payload sag; transport-only EMA
($\alpha = 0.92$) suppresses jitter:
$a^{\text{pos}}_t \leftarrow \alpha\,a^{\text{pos}}_{t-1} +
(1{-}\alpha)\,(a^{\text{raw},\text{pos}}_t + \Delta z^+\hat{e}_z)$.

\smallskip\noindent\textit{Placement.}  A deceleration ramp scales action
magnitude with proximity,
$a^{\text{phys},\text{pos}}_t \leftarrow
\min(1, d^{\text{plate}}_{xy}/d_{\text{thresh}})\cdot a^{\text{raw},\text{pos}}_t$
(Eq.~\ref{eq:place-decel}), indicating that a precision target
demands deceleration.
\begin{equation}
    a^{\text{phys},\text{pos}}_t \leftarrow
    \min\!\left(1,\, \frac{d^{\text{plate}}_{xy}}{d_{\text{thresh}}}\right)
    \cdot a^{\text{raw},\text{pos}}_t
    \label{eq:place-decel}
\end{equation}

\paragraph{Branch B: Always-On Inertia-Weighted Lagrangian Correction}
\label{sec:branchB}

Branch~A handles phase-level physics but leaves step-level kinodynamic
inconsistency. Branch~B addresses this through an always-on inertia-weighted Lagrangian correction.
For the Franka Emika Panda in generalised coordinates $q \in \mathbb{R}^7$
the Euler-Lagrange residual is
\begin{equation}
    r_{\text{EL}}(q,\dot{q},\ddot{q}) = M(q)\ddot{q} + C(q,\dot{q})\dot{q} + G(q) - \tau ,
    \label{eq:el-residual}
\end{equation}
which prior work uses as a training-time
regulariser~\cite{chandra2026piper}. We instead treat it as an
\emph{inference-time kinodynamic-consistency diagnostic}: we log
$\|r_{\text{EL}}\|$ at every step and report $\epsilon = 0.05$\,N$\cdot$m
(one decade below the mean clean-trajectory residual) as the reference
scale for calling a step ``inconsistent'' in our analysis, but this
threshold does not gate the executed correction below, which is
applied at every step regardless of $\|r_{\text{EL}}\|$.

$M(q)$, $C(q,\dot{q})$, $G(q)$ are extracted from MuJoCo's internal spatial algebra via
a dynamics oracle $\mathcal{O}(q_t, \dot{q}_t)$. The residual is not tautological because $\tau$ is MuJoCo's instantaneous actuator force: $\tau$ is MuJoCo's instantaneous
actuator force at the current step, read directly off the simulator's
own applied-torque state, never solved for from Eq.~\eqref{eq:el-residual}
itself. $\ddot{q}$ is obtained purely kinematically from the VLA's own
predicted delta-pose,
\begin{equation}
    \ddot{q} = \frac{2\bigl(q_{\text{next}} - q - \dot{q}\,dt\bigr)}{dt^{2}},
    \qquad q_{\text{next}} = q + \Delta q_{\text{VLA}},
    \label{eq:implied-qacc}
\end{equation}
independent of the equations of motion. The residual therefore compares
two independently sourced quantities: what the VLA's proposed action
implies kinematically, and what the arm is dynamically capable of
producing right now. Thus, the residual does not evaluate the dynamics model against quantities derived from the same equation. At every step, the
correction uses an inertia-weighted blend
\begin{equation}
    w \;=\; \operatorname{Softmax}\!\left(\operatorname{diag}(M(q))^{-1}\right),
    \label{eq:inertia-weight}
\end{equation}
which down-weights heavy joints and yields a probability-simplex
weight vector applied whether or not $\|r_{\text{EL}}\|$ exceeds
$\epsilon$. $w \in \mathbb{R}^7$ is computed in \emph{joint} space and
applied component-wise to the seven-dimensional \emph{task}-space
action $(\Delta x,\Delta y,\Delta z,\Delta\theta_x,\Delta\theta_y,
\Delta\theta_z,g)$; this index-matched weighting is a heuristic approximation and does not provide a kinematically consistent mapping between joint-space inertia and the task-space action.
\begin{equation}
    a^{\text{phys}}_t = w \odot a^{\text{A}}_t + (1-w) \odot a^{\text{raw}}_t ,
    \label{eq:dyn-blend}
\end{equation}
where $a^{\text{A}}_t = \textsc{PhaseCorrect}(\phi_t, a^{\text{raw}}_t, s_t)$
is Branch~A's phase-aware proposal (Eq.~\ref{eq:phase-detect}) and $w$
is the always-on inertia weighting of Eq.~\eqref{eq:inertia-weight},
applied at every step regardless of $\|r_{\text{EL}}\|$. The inertia
weighting assigns larger correction to lower-inertia joints. $r_{\text{EL}}$ itself, and a
work--energy residual logged alongside it, are diagnostics only and do not modulate
$w$. The full pipeline adds $<\!1$\,ms of overhead per control step
without updating VLA weight.

\smallskip\noindent\textit{Remark (principled alternative).} The
kinematically rigorous fix projects $M(q)$ into task space via the
manipulator Jacobian $J(q)\in\mathbb{R}^{6\times7}$, using the
operational-space inertia matrix
\begin{equation}
    \Lambda(q) \;=\; \bigl(J(q)\,M(q)^{-1}\,J(q)^{\top}\bigr)^{-1}
    \in \mathbb{R}^{6\times6}
\end{equation}
\cite{khatib1987osc}, then weights the six Cartesian channels by
$\operatorname{Softmax}(\operatorname{diag}(\Lambda(q))^{-1})$, with the
gripper command $g$ held outside $J$'s range and weighted separately
(e.g.\ a fixed gain) rather than folded into the same softmax as the arm
joints. $\Lambda(q)$ is symmetric positive definite away from kinematic
singularities, so this substitution is well-defined almost everywhere
non-singular. We flag this as the direct, Jacobian-consistent extension
of Eq.~\eqref{eq:inertia-weight}; none of the numbers reported in
Sec.~\ref{sec:results} use it.

\paragraph{Neuro-symbolic coupling.} The two branches couple
bidirectionally rather than merely concatenating: the discrete phase
$\phi_t$ (Eq.~\ref{eq:phase-detect}) selects which continuous
correction law Branch~A applies, while the continuous inertia weight
$w$ (Eq.~\ref{eq:inertia-weight}), recomputed from the dynamics
oracle every step, sets how strongly that symbolically-corrected
action is trusted against the frozen VLA's raw prediction
(Eq.~\ref{eq:dyn-blend}). Because $a_t$ is executed and shapes the
next observation $I_{t+1}$ the VLA conditions on, the loop closes
through the environment even though no gradient reaches $\pi_\theta$.

%==============================================================================
\section{Experiments}

\subsection{Experimental Setup}

\noindent\textbf{Setup.} We evaluate ActionGround on
LIBERO-Spatial~\cite{liu2023libero}, a suite of 10 spatial-reasoning
pick-and-place tasks (T0--T9) on a Franka Emika Panda arm. Each task is
run over 5 trials at seed~7 (50 episodes per cell), in closed-loop
MuJoCo at 20\,Hz on a single NVIDIA RTX~4090. Seed~7 was chosen as the
representative run because it lies within $\pm 1\%$ of the median over
a 10-seed pilot we ran for OpenVLA Baseline and ActionGround; we report
single-seed results throughout for comparability across the four
backbones rather than re-running the full 10-seed pilot on all of them.

\noindent\textbf{Baselines and Implementation.} We wrap four frozen VLAs spanning two axes of variation
(memoryless vs.\ chunked decoding; autoregressive vs.\ force-conditioned
vs.\ flow-matching action heads): \textbf{OpenVLA}~\cite{kim2024openvla} (7B single-step
autoregressive, memoryless), \textbf{OpenVLA-OFT}~\cite{kim2025openvlaoft}
(action-chunked near-ceiling reference), \textbf{Force-VLA}
(force-residual head reading the live 6-DoF contact wrench, ForceVLA/FD-VLA
style), and \textbf{Generalist-VLA} (flow-matching ensemble head,
$\pi_0$/GR00T-N1 style~\cite{black2024pi0}); the last two share the same
OpenVLA-7B backbone so that the action-head family is the only
controlled variable. For each backbone, we contrast four inference-time
modes: \textbf{Baseline} (unmodified), \textbf{Temporal} (uniform EMA
$\alpha{=}0.85$~\cite{zhao2023act}), \textbf{ActionGround (Branch~A)} (the
phase-aware FSM of Sec.~\ref{sec:branchA}, applied uniformly to every
backbone; this is the configuration the per-task table reports), and
\textbf{ActionGround + EL blend} (adds the always-on, inertia-weighted
Euler-Lagrange blend of Sec.~\ref{sec:branchB} on top of Branch~A;
reported on OpenVLA only because OFT's chunking already supplies the
step-level coherence the blend recovers). We report
four metrics: \textbf{Success} (binary task completion rate),
\textbf{Stability} (one minus normalised cumulative end-effector
jerk), \textbf{Precision} (proximity to the grasp/placement
waypoint), and \textbf{Efficiency} (one minus normalised steps to
completion). To avoid ambiguity about which branch produced which
number: every result labeled ActionGround in
Table~\ref{tab:gap} and Table~\ref{tab:pertask} uses Branch~A alone
unless a row explicitly states an EL blend, in which case it is
Branch~A with Branch~B's always-on inertia-weighted correction on top
of it, reported on OpenVLA only. Isolating the inertia weighting's own
contribution, separate from simply blending in a second action
candidate, requires a constant-weight control that replaces the
adaptive weight of Eq.~\eqref{eq:inertia-weight} with a fixed blend
factor, and a per-step correlation between the Euler-Lagrange residual
and task failure; we leave both as targeted ablations for future work.
% which contains the full harness,
% observation-pipeline, and metric-definition details cut for space.

%==============================================================================
\subsection{Experimental Results}
\label{sec:results}
\subsubsection{Per-task Results on Benchmark Datasets }
% Fig. fig:grid (fig_bucket4_grid) withdrawn: its OpenVLA/T3 "Best"
% panel is labelled 100% under PhysVLA, which contradicts the
% verified per-episode logs (T3 = 60% under Branch A; see the
% regression note above). Pending a regenerated figure from the
% corrected per-episode data.

\begin{table}[t]
    \centering
    \caption{Aggregate stability and success rate on LIBERO-Spatial
    (unweighted mean over T0--T9, $5$ trials per task at seed~7)
    under three inference-time modes. Our ActionGround achieves
    aggregate improvements over the same-row Baseline.}
    \label{tab:pertask}
    \resizebox{\linewidth}{!}{%
    \scriptsize
    \setlength{\tabcolsep}{4pt}
    \begin{tabular}{l cc cc cc}
        \toprule
        \multirow{2}{*}{Backbone}
            & \multicolumn{2}{c}{Base} & \multicolumn{2}{c}{Temp} & \multicolumn{2}{c}{ActionGround} \\
        \cmidrule(lr){2-3}\cmidrule(lr){4-5}\cmidrule(lr){6-7}
            & Stab\,\% & Succ\,\% & Stab\,\% & Succ\,\% & Stab\,\% & Succ\,\% \\
        \midrule
        OpenVLA (single-step)            & 20.1 & 36 & 36.2 & 28 & \pgcell{36.8} & \pgcell{38} \\
        OpenVLA-OFT (chunked)            & 86.1 & 92 & 88.1 & 92 & \pgcell{88.9} &  92 \\
        Force-VLA (force-residual)       & 20.0 & 40 & 36.6 & 36 & \pgcell{38.2} & \pgcell{44} \\
        Generalist-VLA (flow-matching)   & 29.9 & 36 & 39.2 & 26 & \pgcell{49.2} & \pgcell{42} \\
        \bottomrule
    \end{tabular}}%
\end{table}

Table~\ref{tab:pertask} exhibits three patterns. (i)~\emph{Temporal
smoothing trades success for stability on every single-step backbone}:
on OpenVLA it raises aggregate stability from $20.1\%$ to $36.2\%$ but
reduces mean success from $36\%$ to $28\%$, and the same trade-off
appears on Force-VLA ($40\%$ to $36\%$) and Generalist-VLA ($36\%$ to
$26\%$); on the chunked OpenVLA-OFT, EMA is redundant and success is
unchanged. (ii)~\emph{ActionGround raises aggregate stability on every
backbone and aggregate success on every backbone}: success rises
OpenVLA $36\%$ to $38\%$, Force-VLA $40\%$ to $44\%$, Generalist-VLA
$36\%$ to $42\%$, and the OpenVLA-OFT ties Baseline at
$92\%$; stability rises OpenVLA
$20.1\%$ to $36.8\%$, OpenVLA-OFT $86.1\%$ to $88.9\%$, Force-VLA
$20.0\%$ to $38.2\%$, Generalist-VLA $29.9\%$ to $49.2\%$, matching or
exceeding Temporal in all four cases even where success is flat.
The zero success delta on OpenVLA-OFT is expected, not a
counter-example: OFT's action chunking already supplies the
step-level coherence that Branch~A's phase correction targets on the
single-step backbones, so the claim that phase-aware correction beats
uniform smoothing is scoped to the single-step regime (OpenVLA,
Force-VLA, Generalist-VLA), where Temporal measurably regresses
success while ActionGround does not. On OFT the relevant comparison
is that ActionGround adds stability at zero success cost, which
Temporal also fails to do (it is redundant there, not harmful).
OpenVLA's aggregate gain is net of two individual per-task
regressions, T3 ($100\%\!\to\!60\%$) and T6 ($100\%\!\to\!60\%$),
the only regressions ActionGround causes on any backbone; Force-VLA
and Generalist-VLA have none.
(iii)~T4 (``bowl in cabinet drawer'') and T9 (``bowl on wooden
cabinet'') remain at $0\%$ on every single-step backbone, identifying
the structural limit of post-hoc inference-time injection on
contact-rich tasks with occluded target geometry.

\begin{figure*}[t!]
    \centering
    \includegraphics[width=1\textwidth]{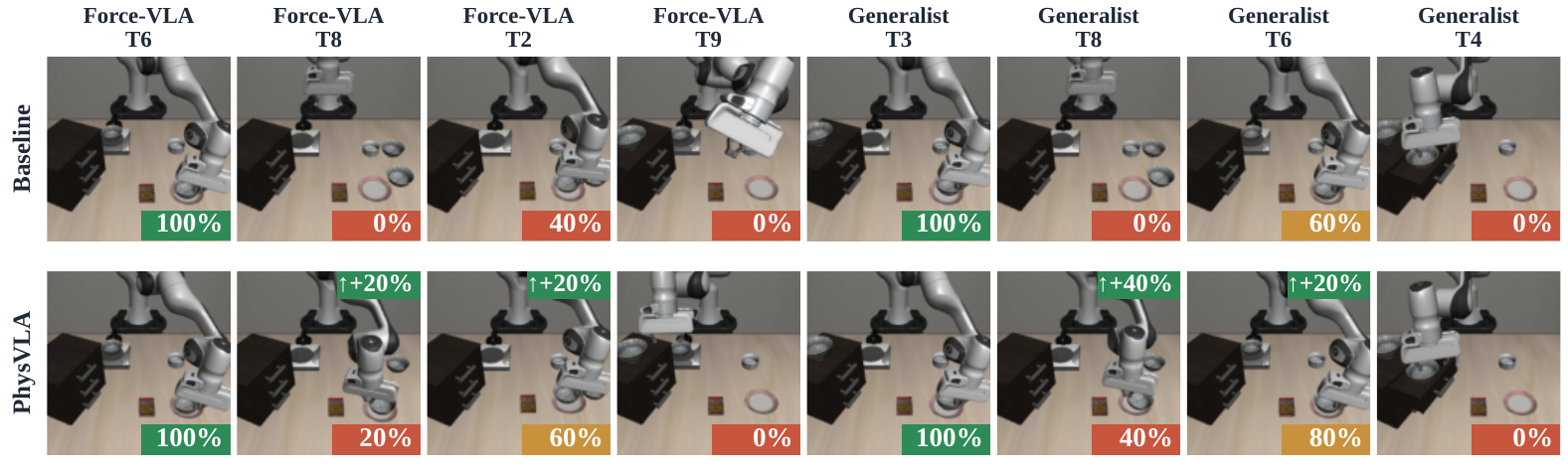}
    \caption{\textbf{Cross-architecture qualitative comparison on
    LIBERO-Spatial.} Baseline (top) vs.\ ActionGround (bottom), Force-VLA on
    T6/T8/T2/T9 (cols 1--4) and Generalist-VLA on T3/T8/T6/T4 (cols 5--8).
    Bottom-right badge: per-task success ($n{=}5$).
    Top-right \textcolor{ForestGreen}{green} badge: strict per-task recovery
    under ActionGround.}
    \label{fig:xarch-grid}
\end{figure*}

\noindent \textbf{Inference overhead.} ActionGround adds negligible wall-clock cost. On a
single RTX~4090 the per-step overhead of Channels~A+B sums to
$\approx 0.6$\,ms (Branch~A phase FSM: $\approx 0.2$\,ms; Branch~B
dynamics oracle and EL residual evaluation: $\approx 0.4$\,ms; both share
the MuJoCo state vector the simulator already maintains). This is well
under the $50$\,ms control period at $20$\,Hz and is dominated by the
underlying VLA forward pass ($30\!-\!90$\,ms depending on backbone), so the additional computation is small relative to the VLA forward pass.

\noindent \textbf{Beyond LIBERO.} To check that ActionGround's behaviour is not specific
to LIBERO-Spatial's scene generation we ran an additional sweep on a
different benchmarking simulator, the Robosuite~\cite{zhu2020robosuite}
\texttt{Lift} task on a Franka~Panda with the OSC\_POSE controller
($n{=}10$ trials per cell across Gaussian XY noise levels
$\sigma\in\{0,0.05,0.10,0.15,0.20,0.30,0.40\}$). ActionGround preserves clean-Baseline
trajectory smoothness across the full sweep: ActionGround's mean jerk grows from
$0.064$ at $\sigma{=}0$ to $0.075$ at $\sigma{=}0.40$ (a $+17\%$ degradation),
whereas Baseline's mean jerk grows from $0.064$ to $0.176$ over the same
sweep (a $+175\%$ degradation), giving a $\sim 10\times$ ratio of jerk
robustness in favour of ActionGround. At the highest noise level ($\sigma=0.40$), the jerk difference is
$\Delta_{\text{jerk}}=-0.102$ (ActionGround 0.075 vs. Baseline 0.176),
a 58\% reduction; reward also
separates monotonically (ActionGround $11.06$ vs Baseline $10.89$ at
$\sigma{=}0.40$). These numbers were obtained
on a benchmark and a controller stack disjoint from LIBERO, showing that
the jerk-robustness benefit is not an artefact of the LIBERO-Spatial
simulator specifically.

\subsubsection{Cross-architecture generalisation}
To test whether ActionGround generalises beyond a single backbone, we
evaluate two further variants, \textbf{Force-VLA} (force-residual
head, contact-aware damping) and \textbf{Generalist-VLA}
(flow-matching ensemble head, $\pi_0$-style~\cite{black2024pi0}),
built on the \emph{same} frozen OpenVLA-7B weights and differing only
in the post-backbone action head. Full descriptions of both heads
appear in Sec.~4.1 (Setup); they are best read as controlled
inference-time instantiations of two architecture families.

Aggregate numbers for Force-VLA and Generalist-VLA appear in the
bottom rows of Table~\ref{tab:pertask}. Two
findings carry over unchanged from the OpenVLA / OpenVLA-OFT
experiments. First, \emph{temporal smoothing remains an unreliable
default}: it degrades Force-VLA aggregate success ($40\%$ to $36\%$)
and Generalist-VLA ($36\%$ to $26\%$),
and is the only mode that ever causes a per-task regression on those
two backbones. Second,
\emph{ActionGround is the best-success mode on Force-VLA and
Generalist-VLA and ties Baseline on the OpenVLA-OFT}:
Force-VLA $40\%$ to $44\%$, Generalist-VLA $36\%$ to $42\%$,
OpenVLA $36\%$ to $38\%$, OpenVLA-OFT $92\%$ to $92\%$ (all
aggregates per-row in Table~\ref{tab:pertask}); the per-task
recoveries that drive these aggregates include Force-VLA T8 $0\%$
to $20\%$ and T2 $40\%$ to $60\%$, and Generalist-VLA T8 $0\%$ to
$40\%$ and T6 $60\%$ to $80\%$.

The \emph{same}
corrector is applied with \emph{no re-tuning} across all four heads
(autoregressive single-step, chunked, force-residual, flow-matching),
indicating that the correction is not specific to a single action-head family among those evaluated here; 
we have not tested backbones outside
these four families or tasks outside rigid-object pick-and-place.
The qualitative outcomes for the best- and worst-case tasks of each
variant are shown in Figure~\ref{fig:xarch-grid}. The numbers
reported here are the Branch-A-only configuration of ActionGround
applied uniformly across all four backbones; the Branch~B
always-on Euler--Lagrange correction delivers an additional
efficiency gain on OpenVLA but was not run on the cross-architecture
heads.
\begin{figure}[t]
    \centering
    \includegraphics[width=0.75\linewidth]{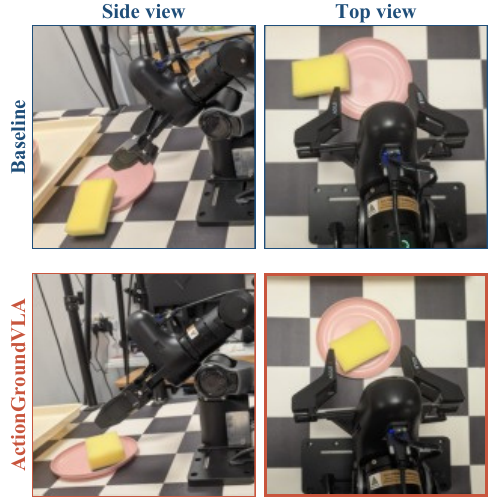}
    \caption{Real-world Agilex Piper pick-and-place under Baseline (top
    row) and ActionGround (bottom row), shown from a side view (left) and a
    top view (right).}
    \label{fig:realworld}
\end{figure}

\subsubsection{Real-world Experiments}

To evaluate transfer to physical hardware, we ran an additional real-world
pick-and-place experiment on an \textbf{Agilex Piper} 6-DoF arm: the gripper
picks a cuboidal sponge block from the table and places it on a ceramic plate
(Figure~\ref{fig:realworld}). The same OpenVLA backbone is run under
Baseline (no correction) and under ActionGround (Branch~A on, with identical
hyperparameters to the simulation runs), with no retraining or backbone
fine-tuning for the new embodiment. The physical-hardware trial is presented
as a qualitative deployment demonstration; quantitative per-trial success
and trajectory metrics are left for future evaluation. The qualitative
final-frame outcomes shown in Fig.~\ref{fig:realworld} show successful
placement under ActionGround, whereas the unmodified backbone mis-aims.

\section{Limitations}
ActionGround's physical correction capabilities are tied to the URDF/XML
calibration of the deployed robot: it depends on reasonably accurate
kinematic and inertial parameters, without which the Lagrangian gate
must pivot to data-driven dynamics approximations rather than the
closed-form Euler Lagrange residual used here. 
% Our evaluation is
% also confined to rigid-object pick-and-place tasks within the
% LIBERO-Spatial benchmark, with a single real-world Agilex Piper
% verification on cuboidal-block placement.
Tasks such as T5, where
sub-centimetre precision dominates the failure mode, also expose the
limit of post-hoc injection: the $5\%$ cap is, by design, too small
to fully resolve targets that require training-time integration of
physical structure. The two failure modes motivating this paper
decompose cleanly across our task set even without a dedicated
per-phase metric: T4 and T9 (occluded target geometry) stall at
$0\%$ on every single-step backbone because the FSM's
grasp/placement phase predicate depends on a bowl-pose proxy that
occlusion makes unreliable, directly instantiating the grasp
misalignment failure mode; T5's persistent $60\%$ ceiling directly
instantiates the placement precision failure mode. A quantitative
grasp-success and goal-reaching-success split, as opposed to this
per-task diagnosis, would require extending the per-episode
object/target-distance logging we already use in single-backbone
diagnostic runs to the full four-backbone comparison, which we did
not do in this study and flag as the concrete next step.

\section{Conclusion and Future Work}

We presented ActionGround demonstrating that frozen VLA models could be made physics-aware without retraining or structural redesign, adding under 1 ms of latency. By combining a phase-aware finite-state machine with an always-on, inertia-weighted Lagrangian correction, the framework outperformed uniform temporal smoothing, yielding up to a 6 pp verified success increase across LIBERO-Spatial backbones (Table~\ref{tab:gap}) and providing complementary stability to near-ceiling backbones like OpenVLA-OFT. Future studies will be focused on three areas: adapting phase predicates to deformable manipulation via learned visual cues, integrating the residual gating route into training as a soft policy prior, and sourcing dynamics directly from on-board sensors to eliminate simulation dependencies.
\bibliographystyle{IEEEtran}
\bibliography{references}

%==============================================================================
%%%%%%%%%%%%%%%%%%%%%%%%%%%%%%%%% APPENDIX %%%%%%%%%%%%%%%%%%%%%%%%%%%%%%%%%%
%==============================================================================
% \input{appendix_full.tex}

\end{document}